# Du TAL au TIL

Michael Zock[1]  Guy Lapalme[2]

(1) Equipe TALEP, LIF, CNRS, UMR 6166
Case 901 - 163 Avenue de Luminy
F-13288 Marseille Cedex 9

(2) RALI-DIRO, Université de Montréal,
CP 6128, Succ. Centre-Ville,
Montréal, Québec, Canada H3C 3J7

> So we need to be alert. It's not just that we may find ourselves putting the cart before the horse. We can get obsessed with the wheels, and finish up with uncritically reinvented, or square, or over-refined or otherwise unsatisfactory wheels, or even just unicycles.
>
> Karen Spärck Jones

**Résumé** papier de *prise de position*.

Historiquement deux types de traitement de la langue ont été étudiés: le traitement par le cerveau (approche psycholinguistique) et le traitement par la machine (approche TAL). Nous pensons qu'il y a une place pour un troisième type: le traitement *interactif* de la langue (TIL), l'ordinateur assistant le cerveau. Ceci correspond à un besoin réel dans la mesure où les gens n'ont souvent que des connaissances partielles par rapport au problème à résoudre. Le but du TIL est de construire des ponts entre ces connaissances momentanées d'un utilisateur et la solution recherchée. À l'aide de quelques exemples, nous essayons de montrer que ceci est non seulement faisable et souhaitable, mais également d'un coût très raisonnable.

**Abstract** position paper.

Historically two types of NLP have been investigated: fully automated processing of language by machines (NLP) and autonomous processing of natural language by people, i.e. the human brain (psycholinguistics). We believe that there is room and need for another kind, INLP: interactive natural language processing. This intermediate approach starts from peoples' needs, trying to bridge the gap between their actual knowledge and a given goal. Given the fact that peoples' knowledge is variable and often incomplete, the aim is to build bridges linking a given knowledge state to a given goal. We present some examples, trying to show that this goal is worth pursuing, achievable and at a reasonable cost.

**Mots-clés :** traitement interactif de la langue, prise en compte de l'usager, outils de traitement de la langue, apprentissage des langues, dictionnaires, livres de phrases, concordanciers, traduction.

MICHAEL ZOCK, GUY LAPALME

# 1   Besoins écologiques d'une discipline, l'exemple TAL

Une discipline doit régulièrement se remettre en cause pour continuer non seulement à avancer, mais surtout dans la bonne direction. Ceci suppose une réflexion concernant les objectifs et des moyens.

En effet, on peut se demander quel est le rôle des chercheurs impliqués dans l'élaboration de produits destinés au traitement de la langue. Est-ce que leur mission consiste uniquement à créer des produits, permettant une prise en charge intégrale du traitement, donc, sans aucune intervention d'un être humain (*traitement automatique*, TAL), ou est-ce qu'ils ne devraient pas également chercher à construire des outils permettant à l'être humain de passer d'une solution partielle (connaissance disponible à un moment donné) à la solution complète ? [1] Il s'agirait donc d'une prise en charge partielle (*traitement interactif*, TIL), on construirait donc des ponts : la machine venant à l'aide de l'être humain.

Voici une question méritant qu'on s'y attarde un peu, car, nombreuses sont les situations où l'aide serait bien utile (qu'on soit en langue maternelle ou en langue étrangère), les performances des êtres humains n'étant pas toujours à la hauteur de leurs intentions. Si les échecs sont relatifs et susceptibles de passer inaperçus, ils sont néanmoins réels, allant du silence (manque de mot) aux erreurs grossières, en passant par des productions approximatives.

L'expérience de correcteurs de fautes d'orthographe est révélatrice à cet égard: d'un côté, les utilisateurs se trouvent submergés/noyés de suggestions non pertinentes, d'un autre côté ils ne trouvent pas la réponse aux questions qu'ils se posent. D'ailleurs, ceux qui ont proposé de faire de la traduction interactive dans les années 70 se sont bien rendu compte qu'il était souvent plus difficile de poser les bonnes questions que d'y répondre ou de traduire le texte lui-même.

C'est donc dire que le dilemme est loin d'être résolu et nous aimerions ici amener quelques pistes de réflexion sur les aspects suivants :
- que le TIL est un secteur important, pas assez exploré, correspondant à un véritable besoin et à un marché ;
- que malgré la qualité de certains programmes ou ressources, ceux-ci peuvent être améliorés en prenant davantage en compte l'utilisateur final ;
- que l'intérêt des ressources augmente avec leur intégrabilité ;
- qu'un moyen d'y arriver serait de partager les ressources dans un format commun.

Afin d'illustrer les deux premiers points, nous allons montrer quelques exemples, programmes existants ou en-cours d'élaboration, afin d'assister l'apprentissage des langues, la rédaction, la lecture ou l'accès aux mots. Notre objectif est moins de convaincre le lecteur de la qualité des solutions envisagées, que de montrer qu'il y a des besoins et, qu'au prix d'efforts raisonnables, on pourrait faire bien plus qu'on ne fait actuellement pour assister l'être humain. Ceci suppose, néanmoins qu'on tienne compte du facteur humain et qu'on l'inclue dans le développement. Malgré le titre, l'idée n'est pas de faire du TIL à partir du TAL, mais plutôt de créer des ponts permettant à la personne en difficulté de passer d'un point A (connaissances momentanément disponibles) au point B (solution finale). Un exemple typique est l'accès lexical. Aussi, il ne s'agit pas d'étudier l'apport du TAL pour l'apprentissage des langues, puisque celui-ci demande selon nous une approche du type TIL. Enfin, le TIL n'est pas une alternative au TAL, mais plutôt un autre axe de recherche, supplémentaire, semblable, tout en étant différent puisqu'il n'est pas entièrement automatique et qu'il part (ou permet d'intégrer) des connaissances de l'homme, connaissances qui sont variables. Ceci dit, les compétences des chercheurs TAL peuvent s'avérer fort précieuses pour faire du TIL. C'est d'ailleurs pour cette raison que nous nous adressons à cette communauté.

---

[1] On notera, que Vauquois qui avait introduit le terme de TAL en 1969 (voir Cori et Léon, 2002) souhaitait distinguer deux types de recherche : (a) celui des linguistes utilisant l'ordinateur comme outil pour mettre en place et tester des théories, outils ou ressources (grammaires, dictionnaires, concordanciers, etc.) ; (b) celui d'équipes pluridisciplinaires s'intéressant aux applications. Celles-ci comprenaient la communication homme-machine, l'enseignement programmé, la traduction automatique, etc.

## 2   Élargissement du cadre en TIL

Un des exemples les plus spectaculaires illustrant qu'on peut faire des choses utiles est le moteur de recherche de Google. Ce programme est capable de nous fournir presque instantanément une longue liste de candidats, contenant souvent la réponse à notre requête. Ce qui est frappant c'est que le programme possède peu de connaissances linguistiques, qu'il est très rapide, et qu'il nous rend souvent d'énormes services. C'est clairement un programme TIL. En même temps, ce programme pose un certain nombre de problèmes : il ramène trop d'informations, les bribes d'informations ne sont pas forcément pertinentes ou dans le bon ordre; il faut admettre que l'usager ne fournit pas beaucoup d'indices sur l'ordre des réponses souhaitées. N'ayant pas une trace des buts censés être atteints, on n'a qu'une liste plate d'URLs visités, il y a un réel danger réel d'oublier ce qu'on cherchait initialement. On part dans une mauvaise direction et au lieu de s'approcher du but, on s'en éloigne sans s'en rendre compte.

Conçu dans une optique TIL, WordNet (WN) a connu davantage de succès auprès des linguistes informaticiens (monde TAL) qu'auprès du grand public (utilisateurs de dictionnaires) et des psychologues. On notera d'ailleurs que les deux mondes s'ignorent. Les psychologues ne citent pour ainsi dire jamais WN (Altmann, 1997), et la communauté WN fait l'impasse sur les travaux des psychologues travaillant sur le dictionnaire mental (Miller, 1990). D'autre part, si WN était davantage développé dans une optique TIL les points suivants seraient davantage problématisés : l'organisation du dictionnaire (structure des différentes catégories lexicales) et la partie "sens", ou les éléments de sens des mots. En absence de définition systématique, on constate qu'une recherche sur Google donne parfois des meilleurs résultats qu'une recherche dans WN. C'est notamment le cas lorsque la définition est absente dans la base de WN.

Le TIL correspond à un besoin réel, car, si les gens ont pratiquement toujours quelques connaissances par rapport au problème à résoudre, il leur manque souvent quelque chose, pour y parvenir. Il faut donc leur fournir l(es) élément(s) manquant(s). Autrement dit, il faut créer des ponts liant les connaissances actuelles d'un utilisateur au but recherché. Le TIL présente donc à la fois un défi et une occasion. C'est un défi, car il exige la prise en compte du facteur humain, aspect trop souvent ignoré en TAL puisque non directement pertinent. Or c'est aussi une belle occasion de réaliser des choses fort utiles, car, contrairement à la machine, on peut s'appuyer sur les connaissances et l'intelligence de l'être humain. Le coût pour relever ce défi est donc raisonnable.

## 3   État des lieux

Si les débuts du TAL ont été marqués par la tentative de construire des systèmes complets (Sabah, 1989 et 1990), on a changé d'approche depuis (Jurafsky, D. and J. H. Martin, 2008 ; Manning et Schuetze, 1999), en favorisant la construction de composants de ressources (outils, bases de données). En effet, il y a de nombreux sites contenant des outils (programmes) ou des données (corpus, dictionnaire).

Beaucoup de choses ont été réalisées et souvent de bonne facture. On peut donc se demander ce qu'il nous faut de plus. Bien sûr, il y a des problèmes qualitatifs et quantitatifs (couverture, adéquation et pertinence du corpus), des problèmes d'intégration et de combinaison.

Mais nous aimerions insister sur les aspects suivants :
- si les outils existants sont bien adaptés au TAL, ils ne le sont pas toujours au TIL ;[2]
- on peut faire des choses fort utiles avec des moyens relativement simples;
- on doit être plus sensible aux besoins des gens et les intégrer dès le début dans le cycle du développement.

## 4   Quelques exemples de TIL

Une langue peut s'apprendre suite à un effort délibéré. On étudie la langue et on fait des exercices (apprentissage volontaire). Ceci dit, on peut aussi l'apprendre de manière involontaire (apprentissage incident). C'est un effet de bord qui se produit pour peu qu'on se livre à des activités comme, la lecture, l'écriture, les jeux, etc. On apprend en parlant et en écrivant tout en commettant des fautes dont la majorité sont d'ailleurs souvent sans conséquence sur la compréhension du message. Les enfants étant joueurs de nature ont moins de scrupules à cet égard que les adultes, ils apprennent généralement les langues plus facilement. Or la possibilité d'interaction et la patience infinie d'un ordinateur peuvent être des atouts précieux pour l'apprentissage des langues malgré le fait qu'il ne soit pas un très bon interlocuteur à cause de sa compréhension limitée.

En limitant la portée du discours et le style d'interaction, il devient possible d'assister intelligemment le locuteur, comme le montrent les quelques exemples suivants.

### 4.1   Les livres de phrases

Nous présenterons trois exemples de livres de phrases développés dans un état d'esprit différent, mais tous intéressants par rapport à notre propos. Leur conception repose en grande partie sur des techniques relativement simples.

Tatoeba [TATOEBA] est une plateforme collaborative visant à rassembler une grande quantité de phrases traduites dans plusieurs langues, dont le japonais. Étant un corpus aligné multilingue, Tatoeba peut assister l'apprentissage des langues via sa mémoire de traduction. L'utilité de cette ressource dépend de la couverture, de la qualité et de l'adéquation des exemples par rapport aux besoins de l'utilisateur, mais comme il s'agit d'une ressource construite de manière collaborative, cela devrait évoluer dans le bon sens. Tatoeba intègre un translittérateur permettant d'annoter les idéogrammes en termes de kana ou de convertir les kanas (syllabaires phonétiques) en lettres romanes.

CIFLI-SurviTra [SURVITRA] est une plate-forme destinée à favoriser l'ingénierie et la mise au point de composants de traduction automatique, à partir d'une mémoire de traduction formée de livres de phrases multilingues avec variables lexicales. C'est aussi un livre de phrases digital multilingue, assistant linguistique pour voyageurs monolingues en situation de "survie". Le corpus d'un domaine, le "Restaurant", a été structuré en sous-domaines avec des phrases alignées et des classes lexicales de locutions quadrilingues (français, hindi, tamoul, anglais). L'approche est générique, donc réutilisable

---

[2] D'ailleurs, même des outils destinés à des êtres humains peuvent poser problèmes. Effectuer une requête dans l'interface du [TLF] qui relève non pas du TAL mais du TIL n'est pas toujours facile. Si l'interface est tolérante vis à vis de l'orthographe (écriture phonétique, erreurs d'accents, omission de tiret) et vis à vis des mots fléchis (écriront -> écrire), les choses deviennent nettement plus compliquées en cas d'une recherche assistée ou complexe. D'ailleurs, on vous prévient : « La pleine maîtrise du formulaire ci-dessous nécessite la lecture des documents : (a) Les recherches complexes, (b) Définition du contenu et (c) Les types d'objets du TLF.»

pour d'autres langues et l'application est susceptible d'être également adaptée pour différents types d'appareils tels les téléphones portables et les assistants numériques.

L'objectif de DrillTutor [DRILLTUTOR] est d'assister des apprenants adultes à acquérir les automatismes nécessaires pour produire à un débit normal les structures fondamentales d'une langue. Le point de départ est une base de données contenant les phrases glanées dans des méthodes de langue ou un livre de phrases destiné à des touristes contenant les expressions de base pour survivre dans des situations courantes : faire des courses, demander un renseignement, etc. Ce genre d'ouvrage contient typiquement les structures fondamentales d'une langue et un lexique de base. Pour aider la navigation et la communication de l'intention de communication les structures sont indexées en termes de but, et certains éléments sont généralisés sous forme de variables lexicales (*je veux une bière* -> <PERSONNE> *vouloir* <BOISSON>) pour permettre à l'utilisateur d'exprimer sa pensée en plusieurs temps : d'abord en termes globaux (but, idée globale), puis successivement en termes lexicaux (instanciation des variables par des valeurs lexicales) et grammaticaux (choix morphologiques concernant le *nombre* et le *temps*).

## 4.2 Aide à la lecture et à la rédaction

S'il y a un domaine qui a énormément changé ces dernières années, c'est bien celui des dictionnaires électroniques. La lexicographie a grandement bénéficié des avancés théoriques et de la puissance des ordinateurs et de la richesse cachée des corpus. Malgré cela on ne tient pas toujours assez bien compte des besoins des utilisateurs.

Trouver des informations dans un dictionnaire n'est pas toujours chose aisée, d'abord cela demande souvent beaucoup de temps, et il peut y avoir des problèmes liés à l'écriture et à la morphologie. Pourtant, les problèmes mentionnés pourraient trouver une solution simple et satisfaisante pour bon nombre de cas. Il suffirait d'ajouter au dictionnaire un translittérateur et un lemmatiseur. En intégrant ce type de fonctionnalité dans des applications courantes (traitement de texte ou navigateur internet), on pourrait désormais lire et comprendre des textes écrits dans une langue étrangère (lecture active). Il suffirait alors de cliquer sur un mot pour voir apparaître un menu "pop up" permettant au lecteur de choisir parmi les informations celle qui l'intéresse à cet instant (traduction, définition, information grammaticale, etc.). La barre d'outils Google [GOOGLE] permet déjà d'afficher la traduction d'un mot anglais dans une page web, mais ne tient pas compte du contexte pour désambigüiser le mot.

Antidote [ANTIDOTE], connu principalement pour son excellent correcteur d'orthographe et de grammaire, donne accès à son dictionnaire et à d'autres outils comme des conjugueurs et la recherche de mots apparentés ou cooccurrents avec une interface intuitive et tolérante. Toutefois, la masse de données et l'interface sont fermées aux autres applications de TAL sauf pour certains langages de script qui ne sont pas courants dans la communauté des chercheurs. Antidote permet un accès au dictionnaire en contexte de lecture ou de rédaction en sélectionnant un menu contextuel, mais ce dernier nous transporte alors dans un environnement de travail initial différent de notre lecture.

ELDIT [ELDIT] illustre qu'un dictionnaire, d'une taille aussi petite que 3000 à 4000 entrées, peut rendre d'énormes services à des apprenants de langue à certaines conditions. Le secret d'ELDIT réside dans le fait que c'est un environnement comprenant une base de données d'environ 400 textes interfacée avec des dictionnaires électroniques allemand et italien. Tous les mots apparaissant dans le texte sont consultables soit via le dictionnaire soit via les textes dans lesquels ils apparaissent. En lançant une requête (on donne un mot dans la fenêtre prévue à cet effet) ou en cliquant sur un mot apparaissant dans un texte on peut avoir accès à tout ce qu'on aimerait savoir à son sujet : sens, traduction, usages/contextes, informations grammaticales, flexions, combinaison avec d'autres mots (collocation), etc. Intégré dans un tel environnement, ce dictionnaire est devenu une véritable aubaine pour l'apprenant.

## 4.3 Textes bilingues

Un concordancier est un programme qui recherche des mots et des expressions dans un corpus de texte. Lorsqu'on soumet une requête, le système fouille sa base de données et affiche toutes les occurrences trouvées, dans leur contexte : la phrase contenant la requête ou un certain nombre de caractères avant et après les mots de la requête. Un concordancier bilingue affichera également la traduction de chaque passage contenant une occurrence dans une langue.

Un concordancier bilingue étant un outil destiné à des rédacteurs, des traducteurs ou des apprenants, il n'est pas encore envisageable d'utiliser des systèmes de traduction automatique pour traduire les passages à la volée. Les concordanciers bilingues qui suivent l'approche TAL, p.ex. [TRANSSEARCH], affichent donc des extraits de textes déjà traduits. L'alignement automatique au niveau de la phrase étant en général assez fiable, il est possible de créer automatiquement des banques de millions de phrases parallèles dans lesquelles le système cherche des occurrences dans une langue et affiche la phrase correspondante dans l'autre. Comme il y a en général plusieurs occurrences d'une même requête et qu'elles n'ont pas toutes été traduites de la même façon, un rédacteur peut ainsi s'inspirer de solutions toutes faites à des problèmes de traductions. Les mémoires de traduction s'inspirent du même principe et sont intégrées dans un environnement de rédaction afin de détecter automatiquement des phrases semblables à des phrases déjà traduites pour en suggérer immédiatement la traduction.

Pour l'apprentissage des langues, il est tout à fait possible d'utiliser cette banque de textes et n'afficher que la langue source. L'apprenant tente de comprendre les phrases et lorsqu'il bute sur une d'entre elles, il pourrait obtenir la traduction, c'est-à-dire la phrase correspondante, par simple clic de souris.

Il existe bien des algorithmes d'alignement de mots utilisés, mais ils ne sont pas encore assez fiables pour identifier les traductions au niveau des mots ou des expressions tout en tenant compte des différentes acceptions pour un même mot. Une autre possibilité qui symbolise plutôt l'approche TIL, utilisée par ELex [ELEX], est de baliser manuellement chaque mot ou expression, comme pour ELDIT mentionné précédemment, en effectuant une désambigüisation; même aidé par des dictionnaires ou des étiqueteurs qui peuvent limiter la gamme des choix, ce travail peut-être assez complexe et fastidieux pour l'enseignant.

## 4.4 Traduction interactive

Depuis la suggestion de M.Kay fin des années soixante (Kay, 1980), plusieurs systèmes de *traduction automatique interactive* ont vu le jour, avec souvent des développements importants. Contrairement aux apparences, le terme n'est pas contradictoire dans le mesure où l'être humain demande (interactivement) de l'aide qui lui est fournit automatiquement par la machine. Pour une étude très complète voir H. Blanchon (2004) qui dans son état de l'art décrit des systèmes interactifs comme TAO de Weidner/CAT, Transactive de ALPS avant de présenter son propre système, LIDIA. A noter qu'à l'époque il n'y avait pas encore la notion de *mémoire de traductions* et que LIDIA intégrait les aspects linguistiques, psychologiques et ergonomiques : le maître du jeu était bel et bien l'être humain (Boitet et Blanchon, 1994).

Une approche un peu différente a été suivie par TransType [TRANSTYPE]. Dans ce projet on propose également un outil d'aide à la traduction misant sur l'interactivité. Le système observe le traducteur qui tape son texte cible et périodiquement propose des extensions à ce texte. Le traducteur peut accepter ces complétions telles quelles, il peut les modifier, ou il peut décider de ne pas en tenir compte en continuant simplement à taper. Après chaque caractère entré ou éliminé par le traducteur, TransType recalcule ses prédictions et en propose de nouvelles. Cette approche au TIL est très prometteuse, mais assez lourde à mettre en place, car elle implique le développement de modèles de traduction adaptés qui doivent être appelés à la volée sur des ordinateurs rapides. Le projet initial visait une clientèle de traducteurs professionnels pour lesquels il a été assez difficile de mesurer des gains en productivité étant donné la nouveauté de l'outil. Or il serait intéressant d'adapter cette approche à des contextes de rédaction bilingue soit par des apprenants ou des rédacteurs qui doivent occasionnellement traduire leurs propres textes. D'autant plus que les usagers deviennent de plus en plus familiers avec des suggestions de complétions offertes par des systèmes informatiques tels qu'on les retrouve lors du remplissage de formulaires web ou de la frappe sur des appareils à clavier limités.

Dans les exemples ci-dessus il était souvent question d'apprentissage. Qu'on ne s'y méprenne pas, ce qui vaut pour l'apprentissage d'une langue étrangère vaut souvent également pour la langue maternelle. On ne possède jamais entièrement une langue, et même en langue maternelle on opère souvent sur la base d'informations incomplètes. Par exemple, on cherche ses mots, d'où l'intérêt d'un dictionnaire mettant à la portée de la souris l'information convoitée.

## 5  Perspectives et conclusion

Nous avons argumenté que le TIL correspond à un vrai besoin, car il est important de capitaliser sur les acquis des connaissances disponibles des utilisateurs à un moment donné. Nous avons souligné l'importance de construire des ponts entre les applications et les gens. Ce point nous paraît capital, car, contrairement à beaucoup de systèmes existants notamment dans le domaine de l'EAO nous nous appuyons sur les connaissances disponibles des gens, connaissances susceptibles de varier d'une personne à l'autre et à tout moment. C'est le système qui s'adapte aux besoins de l'utilisateur. Le chemin proposé (ou possible) pour arriver à la solution doit tenir compte de l'état cognitif du moment. Il ne saurait donc être question de suivre un chemin tracé en avance, une fois pour tout et identique pour tout le monde.

Nous avons identifié quelques outils de TAL qui intègrent déjà ces passerelles et nous avons montré ainsi la possibilité de réaliser des choses très utiles même avec des *moyens du bord*. Des dictionnaires électroniques peuvent, par l'intermédiaire du multifenêtrage et de liens hypertexte, augmenter considérablement la quantité d'information accessible, mais ceci suppose une volonté d'intégration dès leur conception alors qu'ils ont été trop souvent développés comme des alternatives à des versions papier.

On n'apprend pas seulement à l'école et de manière volontaire, on apprend pendant toute notre vie, même notre langue maternelle. On apprend beaucoup de choses tout en travaillant sur autre chose : la lecture s'avérant bénéfique à l'expression orale. L'être humain est toujours intéressé par quelque chose, et en gardant intacte cette envie naturelle, voire pour la stimuler il sera utile de créer des ponts, et c'est ici que les acteurs du monde TAL ont un rôle important à jouer.

Bien que beaucoup d'applications interactives aient été réalisées, il y a de nombreuses possibilités qui n'ont pas encore été saisies. Aussi, si nous avons une bonne compréhension des fondements du TAL (ressources, méthodes), l'équivalent nous manque en TIL. Pour y parvenir il faudrait prendre du recul, ce qui suppose qu'il y ait une communauté s'intéressant à ce type de problèmes. Or s'il y a des chercheurs s'intéressant à l'EAO (enseignement assisté par ordinateur), au e-Learning et à la TAO (traduction assistée par ordinateur), il n'y a pas de communauté véritable s'intéressant au TIL. Aussi, il y a ni ouvrage ni méthode décrivant comment tirer bénéfice des points communs (principes). Pourtant, ceci tout comme l'interdisciplinarité sont des facteurs essentiels si l'on veut vraiment progresser sur cette voie qui nous paraît importante.

# 6 URLS

## 6.1 Outils

| | |
|---|---|
| [ATALA] | http://www.atala.org/-Outils-pour-le-TAL-, Répertoire d'outils pour le TAL, plusieurs pour le traitement du français |
| [NLTK] | http://www.nltk.org/Home, Packages Python pour le TAL |
| [OPENNLP] | http://opennlp.sourceforge.net/projects.html, Organizational center for open source projects related to natural language processing |
| [STATNLP] | http://nlp.stanford.edu/links/statnlp.html, Listes d'outils pour le traitement statistique de langue |

## 6.2 Ressources

| | |
|---|---|
| [BLF] | http://ilt.kuleuven.be/blf/, Base lexicale du français |
| [CNRTL] | http://www.cnrtl.fr/, Centre National des Ressources Textuelles et Lexicales |
| [ELDIT] | http://www.eurac.edu/eldit, Environnement d'apprentissage en allemand et en italien basé sur des textes balisés |
| [EURALEX] | http://www.euralex.org/resources/Resources.html, Répertoire de liens vers de dictionnaires et textes |
| [TLF] | http://atilf.atilf.fr/tlf.htm, Trésor de la langue française informatisé |

## 6.3 Systèmes

| | |
|---|---|
| [ANTIDOTE] | http://www.druide.com, Correcteur de fautes d'orthographe intégré des outils de rédaction (commercial) |
| [GOOGLE] | http://www.google.com/tools/firefox/toolbar/FT2/intl/fr/, Barre d'outils de Google (Firefox ou IE) (gratuit dans un contexte commercial) |
| [TRANSSEARCH] | http://www.tsrali.com, Concordancier bilingue (commercial) |
| [TRANSTYPE] | http://rali.iro.umontreal.ca/Traduction/TransType.fr.html, Prototype de traduction interactive |

## 6.4 Apprentissage des langues

| | |
|---|---|
| [DRILLTUTOR] | http://pageperso.lif.univ-mrs.fr/~michael.zock/DrillTutor, Livre de phrases orientées par les buts |
| [ELEX] | http://tice.det.fundp.ac.be/clarolex/, Démonstration de E-LEX pour l'apprentissage de l'anglais par étiquetage |
| [SURVITRA] | http://aiakide.net/survitra-3/, Livre de phrases de *survie* dans un domaine |
| [TATOEBA] | http://tatoeba.org/fre, Livre de phrases interactif et collaboratif |

# 7 Références